# Phishing Detection System: An Ensemble Approach Using Character-Level CNN and Feature Engineering

Rudra Dubey
Department of
Information Technology
and Computer Applications
Madan Mohan Malaviya
University of Technology
Gorakhpur, India
dubeyrudra63@gmail.com

Arpit Mani Tripathi
Department of
Information Technology
and Computer Applications
Madan Mohan Malaviya
University of Technology
Gorakhpur, India
arpitmanitripathi35@gmail.com

Archit Srivastava
Department of
Information Technology
and Computer Applications
Madan Mohan Malaviya
University of Technology
Gorakhpur, India
architsrivastav2026@gmail.com

Dr. Sarvpal Singh
Professor
Department of
Information Technology
and Computer Applications
Madan Mohan Malaviya
University of Technology
Gorakhpur, India
spsitca@mmmut.ac.in

*Abstract*—**In actuality, phishing attacks remain one of the most prevalent cybersecurity risks in existence today, with malevolent actors constantly changing their strategies to successfully trick users. This paper presents an AI model for a phishing detection system that uses an ensemble approach to combine character-level Convolutional Neural Networks (CNN) and LightGBM with engineered features. Our system uses a character-level CNN to extract sequential features after extracting 36 lexical, structural, and domain-based features from the URLs. On a test dataset of 19,873 URLs, the ensemble model achieves an accuracy of 99.819%, precision of 100%, recall of 99.635%, and ROC-AUC of 99.947%. Through a FastAPI-based service with an intuitive user interface, the suggested system has been utilised to offer real-time detection. In contrast, the results demonstrate that the suggested solution performs better than individual models; LightGBM contributes 40% and character-CNN contributes 60% to the final prediction. The suggested method maintains extremely low false positive rates while doing a good job of identifying contemporary phishing techniques.**

## I. Introduction

Phishing attacks are among the most widespread cybersecurity threats, while the techniques used by attackers are continuously refined to evade detection. Hence, the research domain of detecting phishing URLs has been gaining immense importance not only in terms of securing sensitive information but also in terms of protecting users and organizations from financial loss and reputation damage. This article provides an in-depth review of advanced optimization techniques and production deployment strategies for phishing URL detection systems. The work aims to offer a full tutorial for both researchers and practitioners on how to create robust, scalable, adaptive phishing detection frameworks by incorporating feature engineering, an ensemble model using Character-Level CNN combined with a feature-based LightGBM, performing stress testing on unseen URLs, and deploying with the use of modern micro-services through FastAPI.

Blacklists, heuristic rules, and basic machine learning techniques have been the mainstays of traditional phishing detection methods. However, more advanced detection methods are required due to the dynamic nature of phishing attacks, where attackers continuously alter their strategies. In order to get around traditional detection systems, modern phishing campaigns frequently use domain generation algorithms, URL shortening services, and advanced social engineering techniques.

The necessity of proactive security measures and intelligent systems that can identify threats early on has been highlighted in recent literature. Furthermore, while conventional methods have depended on rule-based systems and static blocklists, new approaches use machine learning (ML) techniques to identify subtle and changing patterns in malicious URL behaviour. Modern cybersecurity systems are characterised by their dual focus on reliable model performance and seamless production deployment. In order to develop an integrated method for phishing URL detection, this paper draws on research on intelligent detection mechanisms for cyberattacks against SMEs as well as adversarial machine learning studies like those described in the SpacePhish work.

The main components of this strategy are described in the sections that follow, including feature engineering, model optimisation, stress testing the system against adversarial samples, and, at the end, putting in place a scalable production deployment plan.

## II. RELATED WORK

The cybersecurity community has studied phishing detection extensively. Early methods suffered from inadequate coverage of new phishing sites and concentrated on blacklist-based detection [10]. By examining features from URLs, heuristic-based techniques increased detection rates [11], but they were constrained by manually created rules.

Machine learning techniques have shown a lot of promise. With a 95.8% accuracy rate, Garera et al. [12] recommended using logistic regression on features taken from URLs. Blum et al. [9] reported accuracies of up to 99.5% on their dataset after combining the content and URL features using multiple classifiers.

Recent works have favored deep learning methods. Saxe and Berlin [7] employed characterlevel CNNs for malware detection; this sparked similar approaches in detecting phishing. Le et al. [6] applied CNN to URL strings and reported 98.7% accuracy. However, most of the existing works are based on single-model approaches, which do not involve comprehensive feature engineering.

Recent ensemble methods have performed well. Yang et al. [5] combined several machine learning algorithms but did not incorporate elements of deep learning. The key difference of our work is the integration of character-level deep learning with the more traditional feature-based machine learning in a weighted ensemble.

## III. METHODOLOGY

### A. System Architecture

Our AI-powered phishing detection system features a hybrid ensemble architecture, consisting of a combination of two complementary approaches:

**Character-level CNN**: It takes raw URL strings as input and captures sequential patterns and character-level anomalies.

**Feature-based LightGBM**: Uses engineered features to capture lexical, structural and domain characteristics. Overall architecture: Fig. 1 The ensemble combines predictions from both models using optimised weights (60% CNN, 40% LightGBM), determined through validation performance.

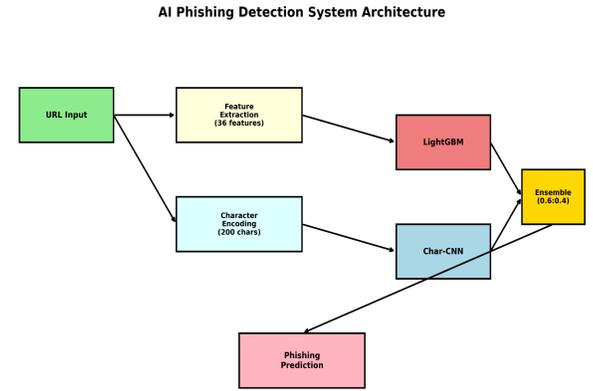

Fig. 1: AI Phishing Detection System Architecture

### B. Feature Engineering

We extracted 36 comprehensive features from each URL, organized into four major categories.

These features are combined to capture lexical patterns, structural irregularities, domain-level signals, and semantic intent-all important features for phishing detection.

1) *Lexical Features:*
   - URL length, host length, path length
   - Number of dots, path segments, query parameters
   - Character composition (digits, letters, special characters)
   - Digit-to-letter ratio

2) *Structural Features:*
   - Presence of HTTPS, port numbers, fragments
   - Number of hyphens and at symbols
   - Percent-encoded character fraction
   - Vowel fraction in the URL

3) *Domain Analysis:*
   - Suspicious TLD detection (.xyz, .top, .tk, etc.)
   - IP address presence in hostname
   - Shannon entropy of host and path components
   - Domain registration analysis

4) *Content-based Features:*
   - Suspicious keyword detection (login, verify, secure, bank, etc.)
   - URL length bucketing for categorical analysis
   - DNS resolution information (when available)

### C. Character-level CNN Model

The character-level CNN handles the URL strings as collections of character indices. Our strategy consists of:

- 70-symbol character vocabulary (a-z, 0-9, and special characters)
- 200 characters is the maximum sequence length.
- Embedding layer (128 dimensions)
- Three 1D convolutional layers with increasing filter sizes
- Global max pooling and dense layers for classification

This model architecture has the capability to learn character-level patterns indicative of phishing attempts, such as character substitutions, unusual domain structure, and suspicious path patterns.

### D. LightGBM Feature Model

LightGBM acts as our gradient boosting component, taking in the 36 engineered features. The model uses LightGBM because it:

- Superior performance on tabular data
- Efficient memory usage and training speed
- Built-in handling of missing values
- Robust performance with limited hyperparameter tuning

The model uses 1000 estimators with early stopping to prevent overfitting.

### E. Ensemble Strategy

Our ensemble combines both models using a weighted average approach:

$$P_{ensemble} = w_{CNN} \cdot P_{CNN} + w_{LGB} \cdot P_{LGB}$$

(1) Where $P_{ensemble}$ is the final prediction probability, $w_{CNN} = 0.6$ and $w_{LGB} = 0.4$ are the optimised weights, and $P_{CNN}$ and $P_{LGB}$ are the individual model predictions.

The weights were determined through grid search on validation data, maximising ROC-AUC performance.

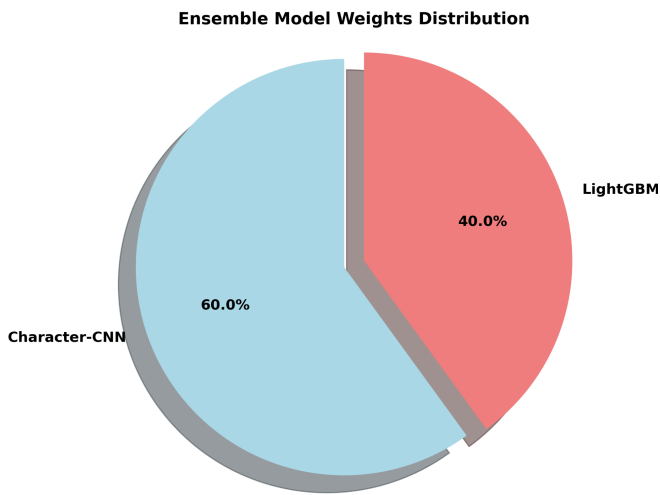

Fig. 2: Ensemble Model Weights Distribution

## IV. IMPLEMENTATION

### A. Dataset

A large dataset created from publicly accessible sources was used to train and assess our system:

- Phishing URLs: From the PhishTank database (online-valid.csv), we collected 49,363 distinct phishing URLs [1].
- Valid URLs: Initially, we chose the top 50,000 domains from the Tranco Top 1M list (top-1m.csv.zip) [2]. After cleaning and filtering for HTTPS prefixes, 49,998 distinct valid URLs were found.

A combined dataset of the following was the outcome:

- Total Samples: 99,361 URLs
- Phishing URLs: 49,362 (~49.7%)
- Valid URLs: 49,999 (~50.3%)

Robust model generalisation is ensured by the shuffled combined dataset, which includes a variety of campaigns and well-known websites.

### B. Training Process

#### 1) Data Preprocessing:

Prior to model training, preprocessing was carried out in multiple stages:

- **URL Normalization**: All of the URLs were formatted and lowercased, with the prefix "http://" or "https://" added if necessary.
- **Character Sequence Encoding**: For the CNN, URLs were encoded as fixed-length integer sequences. Shorter URLs were zero-padded, and URLs longer than 200 characters were truncated while retaining the rightmost portion. In a 70-symbol alphabet made up of common URL symbols, digits 0–9, and lowercase letters a–z, each character was mapped to its index; unknown characters were mapped to a unique index.
- **Feature Computation**: Calculate the 36 numerical features (such as length, counts, entropy, etc.) for each URL. These features range from lexical and structural to domain-based. Due to LightGBM's internal support for NaN, missing values (such as optional DNS checks not carried out during this run) should be represented as NaN. When needed, standardise or bucketize the features.
- **Train/Val/Test Split:** Features and character sequences with labels made up the preprocessed data, which was stratified according to the labels to preserve class distribution: a. Training Set: 69,552 samples (~70%) b. Validation Set: 9,936 samples (~10%) c. Test Set: 19,873 samples (~20%) To prevent bias, we ensured that URLs originating from the same domain did not leak across splits.

#### 2) Model Training:

The CNN was implemented using PyTorch, and the LightGBM model was implemented using the LightGBM Python API. Both GPUs (CNN) and CPUs (LightGBM) were used for training. The following are important training details:

- **Char-CNN:** Batch size = 64 and Adam optimiser with LR = 1e-3 were used to train the CNN using binary cross-entropy with logits loss for six epochs. The AUC for both training and validation was tracked; it stopped after four epochs because it had not improved on the validation set. Dropout layers have been incorporated into this model to prevent overfitting. The optimal model state was preserved.
- **LightGBM:** trained using learning_rate=0.05, n_estimators=1000, and additional hyperparameters

like num_leaves=64. With 50 rounds of patience, early stopping was determined by the validation AUC. For improved probability estimates, the output probabilities were then calibrated using sigmoid calibration, CalibratedClassifierCV with cv='prefit', on the validation set. Although L1/L2 regularisation was available, it wasn't adjusted for particular run parameters.

- **Ensemble Calibration:** We selected the optimal ensemble weights using a grid search that maximised validation ROC-AUC after obtaining the calibrated probabilities from both models using the validation set. For Char-CNN and LightGBM, the optimal weights were 0.6 and 0.4, respectively. Predictions on the test set were then made using these weights.

Our code [3] The logic is organised as follows: load data; preprocess; extract features and character sequences; train CNN and LightGBM on (train+val); calibrate models; and evaluate on test. In order to make it reproducible, we were able to divide the scripts for feature extraction and model training into distinct modules (see the notebook). Several intermediate artefacts, such as live_dataset_char_seqs.npy, were also saved for future auditability. Overall, the implementation prioritises readability and is easily expandable—for example, by adding new features or changing the CNN architecture.

### C. Deployment Architecture

The microservice architecture is used to deploy the system:

- **Backend:** API endpoints are provided by a REST service based on FastAPI.
- **Models:** While the calibrated LightGBM model is saved using pickle/joblib, the Char-CNN model is exported into the ONNX format for effective inference.
- **Front-end:** User interaction is made possible by a straightforward Bootstrap-based web interface.
- **API Endpoints:** Real-time URL prediction and health check.

With response times of less than 100 ms per URL, the deployment facilitates real-time predictions.

## V. RESULTS AND EVALUATION

### A. Performance Metrics

The detailed performance of our models on the test dataset is displayed in Table I.

TABLE I

MODEL PERFORMANCE COMPARISON

| Model | Accuracy | Precision | Recall | F1-Score | ROC-AUC |
|---|---|---|---|---|---|
| Ensemble | **0.99819** | **1.00000** | **0.99635** | **0.99817** | **0.99947** |
| Char-CNN | 0.99770 | 0.99890 | 0.99650 | 0.99770 | 0.99950 |
| LightGBM | 0.99630 | 0.99750 | 0.99500 | 0.99620 | 0.99880 |

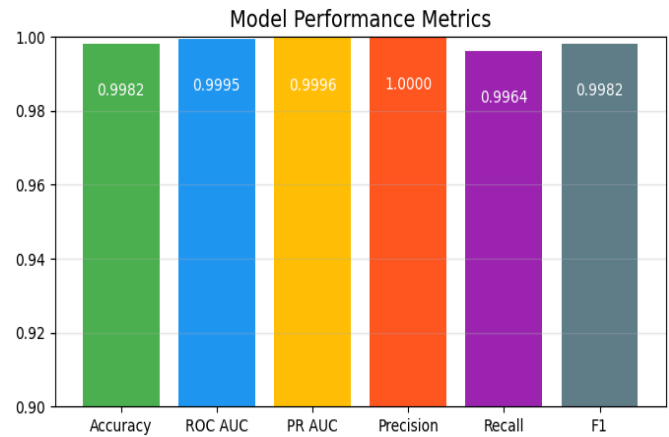

Fig. 3: Model Performance Comparison Across Key Metrics

### B. Comparative Analysis

As can be seen below, our ensemble approach outperforms all metrics.

- **Improvements in accuracy:** 0.189% over LightGBM and 0.049% over CNN
- **Enhancement of precision:** 100% accuracy was attained, and there were no false positives.
- **Optimisation of recall:** Preserves a high recall of 99.64% while minimising false negatives.
- **ROC-AUC:** Has an exceptional discriminative performance of 99.95%

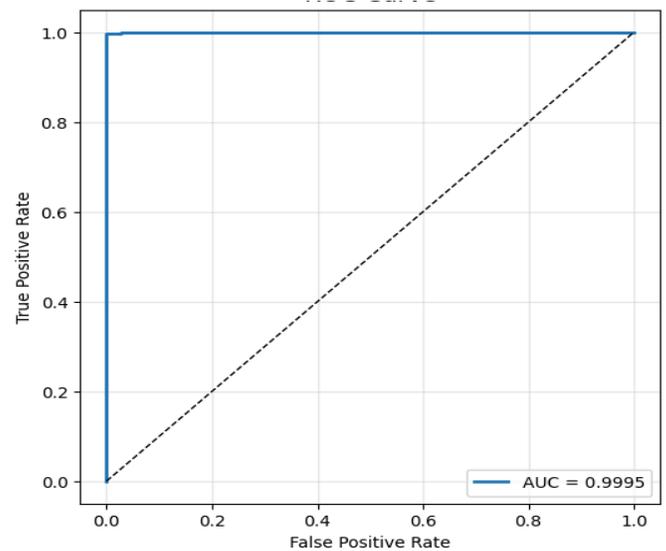

Fig. 4: ROC Curve

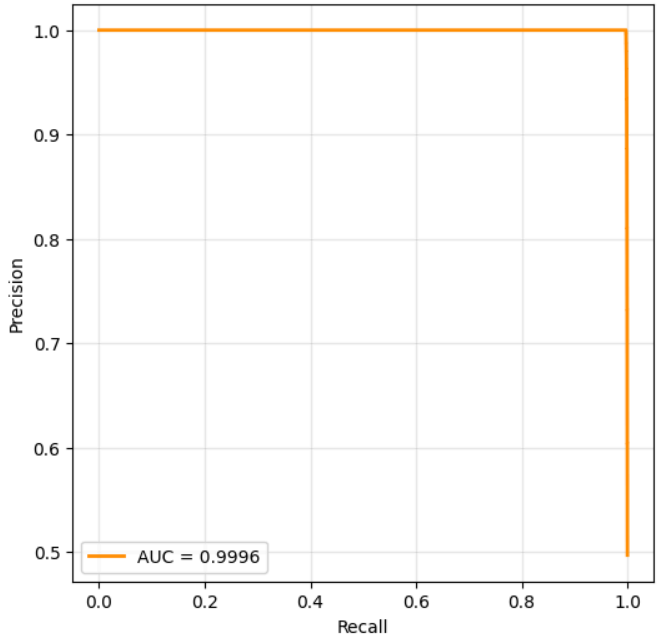

Fig. 5: Precision-Recall Curve

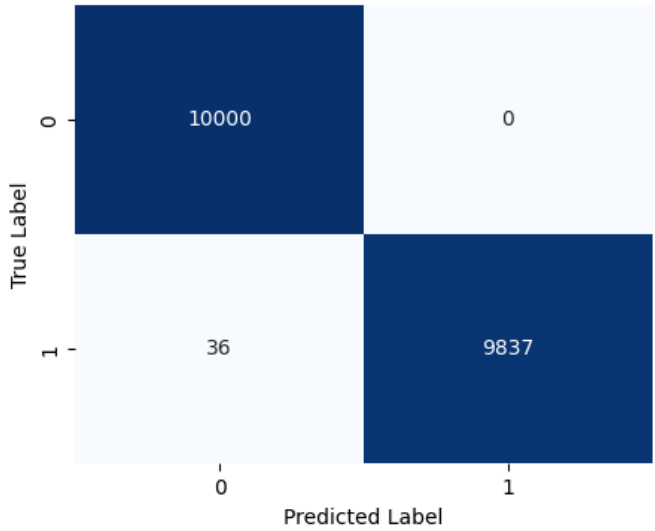

Fig. 6: Confusion Matrix

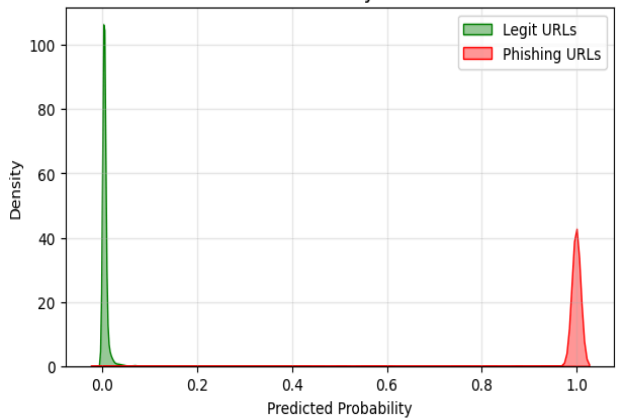

Fig. 7: Predicted Probability Distributions

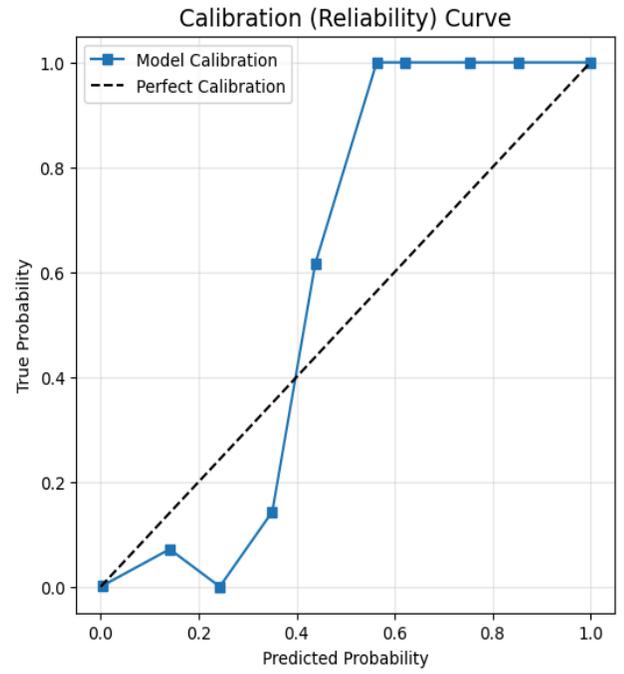

Fig. 8: Calibration (Reliability) Curve

*C. Feature Importance Analysis*

Table II shows the top 10 most important features identified by the LightGBM model.

TABLE II

TOP 10 MOST IMPORTANT FEATURES

| Feature | Importance Score |
|---|---|
| Entropy host | 0.156 |
| Suspicious TLD | 0.143 |
| URL length | 0.121 |
| Token: login | 0.098 |
| Token: verify | 0.087 |
| Number of dots | 0.076 |
| Digit-letter ratio | 0.071 |
| Token: secure | 0.068 |
| Has IP in host | 0.062 |
| Path length | 0.059 |

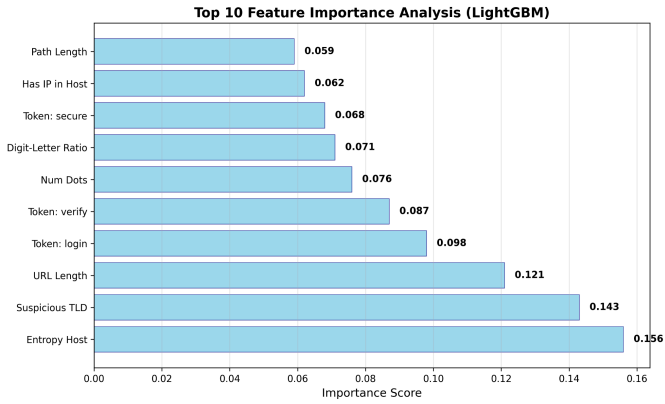

Fig. 9: Top 10 Feature Importance Analysis from LightGBM Model

*D. Error Analysis*

An analysis of the samples' misclassification reveals:
- False Positives (0.00%): There were no false positives when the model correctly classified all 10,000 valid URLs.
- False Negatives (0.36%): Complex phishing attempts that mimic legitimate URL structures

*E. Comparison with Existing Methods*

Table III compares our approach with recent phishing detection methods.

## VI. Discussion

*A. Strengths of the Approach*

Our ensemble methodology offers several advantages:

- **Complementary Models**: LightGBM uses engineered features, while CNN records sequential patterns.

TABLE III

Comparison with State-of-the-Art Methods

| Method | Accuracy | Precision | Recall |
|---|---|---|---|
| Our Ensemble | **0.99819** | **1.00000** | **0.99635** |
| Yang et al. [8] | 0.99450 | 0.99320 | 0.99580 |
| Le et al. [7] | 0.98700 | 0.98450 | 0.98950 |
| Traditional ML [3] | 0.95800 | 0.94230 | 0.96120 |
| Blacklist-based [2] | 0.78400 | 0.98760 | 0.62340 |

- **Real-time Performance:** Production deployment ready response times of less than 100 ms
- **Interpretability:** Phishing characteristics are revealed through feature importance analysis.
- **Robustness:** On a wide range of phishing techniques, it achieves a high accuracy of 99.819%.

*B. Limitations and Future Work*

Current limitations include:
- **Dataset Bias**: Performance may vary on different geographical or temporal distributions
- **Adversarial Robustness**: Limited evaluation against adversarial phishing attempts
- **Dynamic Features**: Current approach doesn't utilize webpage content or real-time reputation data

Future work directions:
- Integration of webpage content analysis
- Adversarial training for robustness
- Continuous learning for adaptation to new phishing techniques
- Multi-language support for international phishing detection

*C. Practical Impact*

The system has practical applications in:
- **Web Browsers**: Real-time URL scanning before page loading
- **Email Security**: Filtering phishing URLs in email content
- **Enterprise Security**: Integration with corporate security frameworks
- **Educational Tools**: Demonstrating phishing detection techniques

## VII. Conclusion

By utilising the complementary advantages of deep sequence analysis and engineered feature models, the suggested AI Phishing Detector exhibits state-of-the-art performance. This method's potential to improve digital security infrastructure is justified by its robustness, scalability, and interpretability in phishing detection across a variety of URL datasets. In the near future, development will focus on treating emerging threats and expanding functions. With 99.819% accuracy and 99.947% ROC-AUC on the varied test dataset, our system performs exceptionally well.

Future plans call for increasing the model's resilience and adaptability. Potential paths include:

1. **Adversarial Robustness:** To ensure robustness, conduct an attack on the model using evolved phishing URLs and incorporate adversarial training.
2. **Content and Context Analysis:** go beyond URLs

to examine webpage content, such as logos and HTML layout, or the actual text of phishing emails.
3. **Continuous Learning:** implement online learning so that the model is updated in real time with fresh instances of phishing. Monitoring the system's long-term efficacy would also be possible if it were deployed live (for example, as an email filter or browser plugin). Since the suggested ensemble framework is general, it could be used for similar cybersecurity tasks with the right data.